\documentclass[conference]{IEEEtran}
\pdfoutput=1   
\IEEEoverridecommandlockouts
\usepackage{cite}
\usepackage{amsmath,amssymb,amsfonts}
\usepackage{graphicx}
\usepackage{textcomp}
\usepackage{xcolor}
\usepackage{booktabs}
\usepackage{multirow}
\usepackage{url}
\usepackage[colorlinks=true,citecolor=blue!60!black,linkcolor=blue!60!black,urlcolor=blue!60!black]{hyperref}
\graphicspath{{figures/}}

\newcommand{\prelimnote}{}
\newcommand{\ragevla}{RageVLA}

\newcommand{\parhead}[1]{\textit{#1:}}

\begin{document}

\title{\ragevla{}: Augmentation-Trained Perception and Action for State-of-the-Art Robustness on LIBERO}

\author{\IEEEauthorblockN{Amrit Lal Singh}
\IEEEauthorblockA{\textit{ragento.ai}\\
\href{mailto:amrit@ragento.ai}{amrit@ragento.ai}}
\and
\IEEEauthorblockN{Smarak Kanjilal}
\IEEEauthorblockA{\textit{ragento.ai}\\
\href{mailto:smarak@ragento.ai}{smarak@ragento.ai}}
}

\maketitle
\pagestyle{plain}\thispagestyle{plain}   

\begin{abstract}
Vision-language-action (VLA) models look saturated on LIBERO, yet most of them fall apart on LIBERO-PRO as soon as an object moves or the instruction asks for something new. We argue that what is missing is not scale but decoupling. In these policies, what to do is tied to where things were during training, to the scene around them and to the wording of the instruction. We cut these ties in two places. In the data, LIBERO demonstrations are re-arranged by general rules: every object and fixture is moved and rotated, tasks are carried into other environments and onto other surfaces, and every instruction is paraphrased. In the policy, a perception engine on frozen DINOv3 features, trained on this augmented data, hands the action model the scene in decoupled form: a list of objects with their positions, and a compact code of what the gripper is looking at, next to image and instruction embeddings from TinyCLIP. A flow-matching diffusion transformer, trained from scratch, acts on this view. We call this combination of \textbf{R}obust \textbf{A}ugmentation and \textbf{G}rounded \textbf{E}ncoding \ragevla{}. With 0.21 billion parameters, \ragevla{} sets a new state of the art on LIBERO-PRO among VLA policies, scoring 0.72 against 0.53 for the best published model, and a larger variant reaches 0.73. Two further tests on LIBERO-Object show that the policy acts on what it perceives. When the object positions it is given are shuffled while the camera images stay correct, its success falls from 0.965 to 0.215. When the target object is displaced by 28~cm, it still succeeds in 0.865 of the episodes, where $\pi_{0.5}$, run through the same harness, succeeds in 0.105.
\end{abstract}


\section{Introduction}
LIBERO~\cite{liu2023libero} has become the default benchmark for vision-language-action (VLA) policies, and its four evaluation suites are now saturated: recent models report average success well above 90\%~\cite{kim2025oft,openpi2025,zheng2025xvla}. LIBERO-PRO~\cite{zhou2025liberopro} asks what these numbers measure. It keeps LIBERO's scenes and perturbs one factor at a time: object appearance (Obj), initial positions (Pos), instruction wording (Sem), the task itself (Task) and the table (Env). Strong LIBERO models collapse under two of these. On the LIBERO-PRO leaderboard, OpenVLA~\cite{kim2024openvla}, $\pi_0$~\cite{black2024pi0}, MolmoAct~\cite{lee2025molmoact} and NORA~\cite{hung2025nora} score zero on almost every Pos and Task cell, and the best model, $\pi_{0.5}$~\cite{pi2025pi05}, averages 0.11 over them. The models have learned to replay trajectories.

\begin{figure}[t]
\centering
\includegraphics[width=\columnwidth]{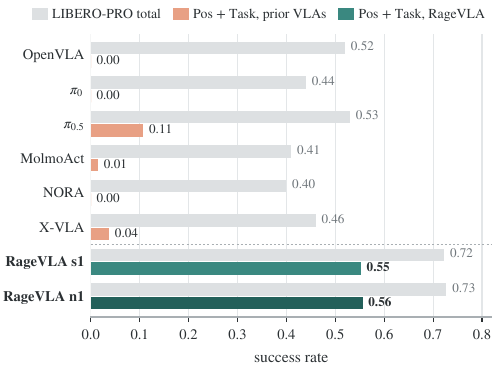}
\caption{LIBERO-PRO's official total (grey) against the mean over the position and task perturbations only (colored). Prior VLAs keep high totals while scoring close to zero once objects move or the task changes. \prelimnote}
\label{fig:teaser}
\end{figure}

The benchmark's headline number hides this. LIBERO-PRO summarizes a model by its mean over all cells, but Obj and Sem leave both the layout and the task unchanged, so a model that replays its training trajectories still scores highly on them. OpenVLA, for example, has a total of 0.52 while solving none of the position-perturbed or task-changed tests. Fig.~\ref{fig:teaser} separates the two readings.

This paper takes a simple position: these policies fail because what they do is \emph{coupled} to things it should not depend on. Each LIBERO task is demonstrated 50 times in almost the same layout, so the identity of an object is tied to where it stands, the motion is tied to the scene around it, and the instruction is tied to one wording and one initial layout. A policy trained on such data has no reason to separate them, whatever its size. Robustness under perturbation requires \emph{decoupling} these factors, and it has to happen in two places at once.

The first is the data. We re-arrange LIBERO's demonstrations by general rules. Re-placing every object and fixture at a random position and orientation decouples \emph{what} an object is from \emph{where} it is. Carrying tasks into other environments and onto other surfaces decouples the manipulation from its background. Paraphrasing decouples the task from its wording. The second is the policy's input. Variation in the data helps only if the policy can read it, so a perception engine hands the action model the same factors in separated form: a list of objects with their classes and global positions, a wrist code of the surfaces and depths in front of the gripper, and one image and one instruction embedding. With accurate object positions as an input, the policy does not need to recall a trajectory; it can act on the scene in front of it. The two parts are designed for each other: without varied data the position input carries no information that the policy needs, and without the position input the variation has to be recovered from pixels. We call the resulting system \ragevla{}, for \emph{Robust Augmentation with Grounded Encoding}.

Our contributions are as follows.
\begin{itemize}
\item \textit{A general augmentation pipeline} for LIBERO-style demonstrations. Every object and fixture is re-placed at a random position and orientation, tasks are transplanted into LIBERO's other environments and onto its other surfaces, and every instruction is paraphrased (Sec.~\ref{sec:aug}). The rules are general: no layout, object pair or instruction is taken from LIBERO-PRO.
\item \textit{\ragevla{}}, an object-centric flow-matching policy. Small perception adapters on a frozen DINOv3 backbone~\cite{simeoni2025dinov3} produce an object list and a wrist-view code, TinyCLIP~\cite{wu2023tinyclip} provides one image and one text embedding, and a diffusion transformer trained from scratch produces actions (Sec.~\ref{sec:model}).
\item \textit{State-of-the-art LIBERO-PRO results among VLA policies} on the full protocol, with 0.21 billion parameters in total (Sec.~\ref{sec:exp}).
\item \textit{Direct evidence that the policy acts on perceived positions.} Changing the object positions given to the policy at inference changes its behaviour even in LIBERO's original layouts, and under LIBERO-PRO's graded object displacement the policy degrades far more slowly than $\pi_{0.5}$ evaluated in the same harness (Sec.~\ref{sec:probe}).
\end{itemize}

\section{Related Work}
\parhead{VLAs on LIBERO} OpenVLA~\cite{kim2024openvla} and OpenVLA-OFT~\cite{kim2025oft} fine-tune a 7B vision-language model. $\pi_0$ and $\pi_{0.5}$~\cite{black2024pi0,pi2025pi05} attach a flow-matching action expert to a 3B vision-language model. MolmoAct~\cite{lee2025molmoact} and NORA~\cite{hung2025nora} decode discrete action tokens autoregressively, and X-VLA~\cite{zheng2025xvla} pairs a Florence encoder with a flow-matching transformer. Smaller policies such as SmolVLA~\cite{shukor2025smolvla} and Octo~\cite{octo2024} trade accuracy for size. All of these compute actions from the image tokens of one vision-language backbone. \ragevla{} factorizes the input instead: explicit object positions and a wrist code come from small perception adapters, and the vision-language input comes from an aligned image-text encoder, TinyCLIP~\cite{wu2023tinyclip}, a distilled CLIP~\cite{radford2021clip}, as one image embedding and one instruction embedding.

\parhead{Robustness benchmarks} LIBERO-PRO~\cite{zhou2025liberopro} and LIBERO-Plus~\cite{fei2025liberoplus} perturb LIBERO's objects, layouts, language, cameras and scenes, and show that LIBERO success does not transfer. We evaluate on LIBERO-PRO with its released files and its own perturbation code, including its graded object-displacement suites, which no leaderboard entry reports.

\parhead{Data generation for imitation learning} MimicGen~\cite{mandlekar2023mimicgen} re-targets object-centric segments of human demonstrations to new object poses and validates them in simulation. Our augmentation follows the same principle on LIBERO and extends it to full-scene re-layouts, fixtures, other environments and other surfaces.

\parhead{Policies and conditioning} Diffusion Policy~\cite{chi2023diffusionpolicy} and flow-matching action heads~\cite{black2024pi0} model action chunks generatively. We use a diffusion transformer~\cite{peebles2023dit} trained with flow matching~\cite{lipman2023flow,liu2023rectified} and the parameter-efficient conditioning of PixArt-$\alpha$~\cite{chen2024pixart}. Our object head follows CenterNet~\cite{zhou2019centernet} on DINOv3 features~\cite{simeoni2025dinov3}.

\parhead{LLM agents as policies} Code-as-policies~\cite{liang2023cap} and, recently, ASPIRE~\cite{aspire2026} use a large language model to write robot programs on top of perception and motion-planning interfaces. ASPIRE reports high success on LIBERO-PRO's position and task perturbations. It is a different paradigm, in which a frontier language model writes and debugs a program for each task, and we discuss it separately from learned VLA policies (Sec.~\ref{sec:aspire}).

\section{Decoupling in the Data: General Augmentation}
\label{sec:aug}
All data starts from LIBERO's human demonstrations, 50 for each of its 130 tasks. Generation stores simulator states and actions only, and images are rendered afterwards. Like MimicGen~\cite{mandlekar2023mimicgen}, we generate new demonstrations by transforming segments of the human ones into new scenes. Each demonstration is cut into object-centric phases (approach, grasp, transport, place, fixture interaction), and each phase is assigned the entity it acts on. For a phase acting on entity $o$, every end-effector pose $T^{W}_{e,t}$ of the source demonstration is first expressed in the frame of $o$ at its source pose $T^{W}_{o}$, and then mapped back to the world through the entity's new pose $\tilde T^{W}_{o}$:
\begin{equation}
\tilde T^{W}_{e,t} = \tilde T^{W}_{o}\,\bigl(T^{W}_{o}\bigr)^{-1}\,T^{W}_{e,t}.
\end{equation}
The relative motion between gripper and object is therefore preserved wherever the object is placed and however it is rotated. Consecutive phases are joined by interpolated transit motions, and the gripper commands are kept from the source. Every generated trajectory is re-simulated and kept only if the task's own success predicate holds at the end. Seeds depend only on the task, the demonstration, the generator and the round, so the data can be regenerated exactly.

\parhead{Random re-layout} In every generated trajectory the scene is laid out anew. In most attempts the layout is uniform: every fixture is placed anywhere on the table with a yaw anywhere in $\pm180^\circ$, carrying its contents, and every free object is placed anywhere within the robot's reach with a yaw drawn according to its symmetry. The others use smaller random displacements of the object to be grasped, the destination and the distractors. Tasks whose instruction is relational (``the bowl \emph{next to} the plate'') receive only the smaller displacements, so that the instruction remains true. Candidate layouts are filtered for workspace bounds, clearance, stability and reachability before re-simulation. No positions are exchanged between objects, and no layout is taken from any file other than LIBERO's own scene definitions.

\parhead{Environment transplants} A task is carried into every other LIBERO environment that can host it, and the same random re-layout is applied on top. A transplant is valid when the task involves at most one fixture and at most three entities in total, and every object or fixture named in the instruction exists exactly once in the target environment, so that the instruction stays unambiguous. Objects are never inserted into a scene. Goal regions that are named after an entity, such as the front of the stove, move with that entity. The transplanted trajectory keeps the instruction of the task it carries.

\parhead{Surface transplants} Every task is also moved onto each of LIBERO's other table surfaces, again with a random re-layout on top.

\parhead{Language} A language model wrote 20 paraphrases for each instruction. During training the original instruction or one of its paraphrases is drawn at random.

\parhead{Quota and split} Each task receives a fixed quota of successful trajectories per round, split between its own environment, other environments and other surfaces (Appendix~\ref{app:impl}). Data is generated in rounds with different seeds; all but one are used for training, and the last is held out as a test set and is used only to measure perception and action error (Table~\ref{tab:data}).

\begin{figure}[t]
\centering
\includegraphics[width=\columnwidth]{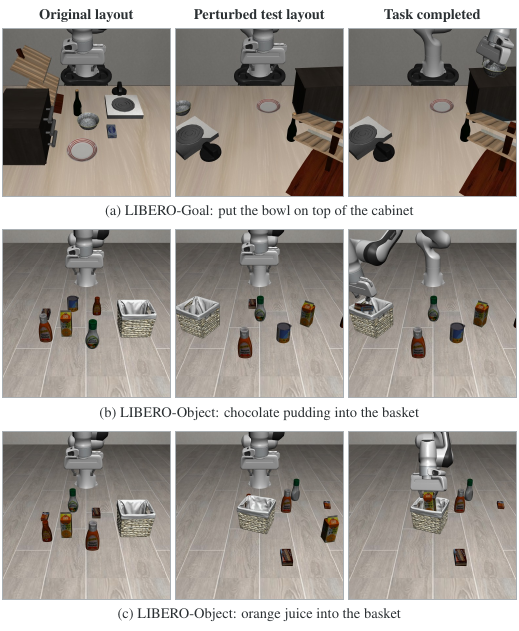}
\caption{\ragevla{} s1 on the held-out test data. Each row is one task: the original LIBERO layout, the perturbed test layout with fixtures and objects moved and rotated, and the scene after the policy has completed the task from the perturbed layout.}
\label{fig:example}
\end{figure}

\parhead{What the data decouples} Each rule removes one correlation that LIBERO's demonstrations contain. Random re-layout decouples the identity of an object from its place: almost no training trajectory keeps an object or a fixture where LIBERO placed it, so the policy cannot succeed by recalling a path and has to read where the objects are from its inputs. Environment and surface transplants, each with a new layout on top, decouple the manipulation from its background: the room and the table change independently of where the objects stand, so neither can serve as a landmark for a stored motion. Paraphrases decouple the task from its wording. The policy thus learns to grasp an object at any point of the workspace, in any orientation, and in any environment or on any surface within LIBERO (Fig.~\ref{fig:example}).

\parhead{General rules, not benchmark cases} None of these rules is derived from LIBERO-PRO's perturbations. LIBERO-PRO's position test exchanges the regions of fixed pairs of objects; we never exchange positions and never read its files, and a swapped layout is simply one of the layouts that uniform placement can produce. Its environment test moves tasks onto one other table (Appendix~\ref{app:protocol}); our rule carries every task into every environment and onto every surface that can host it. We randomize neither object textures nor colors, so appearance is the one factor we leave coupled, and Sec.~\ref{sec:exp} shows where that costs.

\begin{table}[t]
\centering
\caption{Data. All generated trajectories passed the task's success check in simulation. Samples are taken every 8 control steps.}
\label{tab:data}
\footnotesize
\begin{tabular}{lrr}
\toprule
Split & Trajectories & Samples \\
\midrule
Training & 132{,}610 & 2.79M \\
Held-out test & 1{,}751 & 36.6k \\
\bottomrule
\end{tabular}
\end{table}

\section{Decoupling in the Policy: \ragevla{}}
\label{sec:model}
\begin{figure*}[t]
\centering
\includegraphics[width=\textwidth]{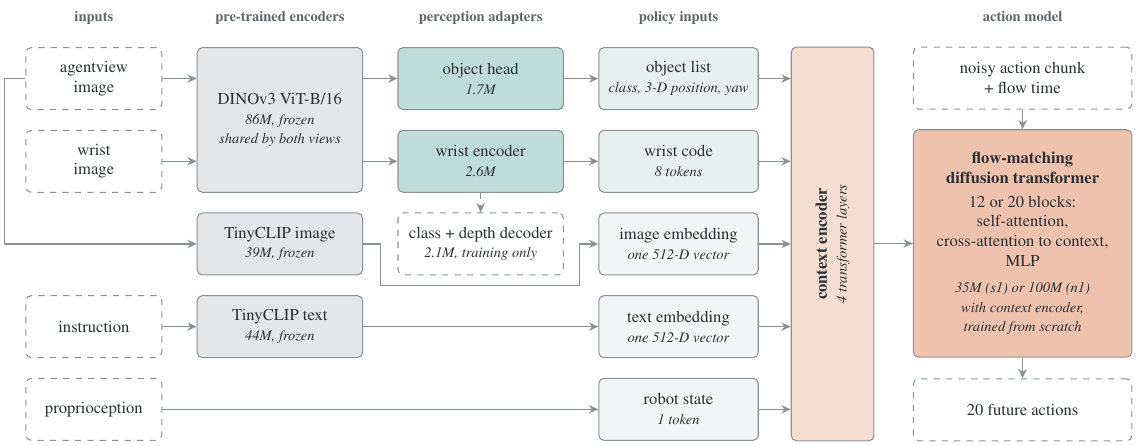}
\caption{\ragevla{}, with the parameter count of every unit. Grey units are pre-trained and frozen, teal units are the trained perception adapters and their outputs, and the action model is shown in coral. Two small adapters on DINOv3 features turn the agentview into an object list and the wrist view into a compact code. TinyCLIP turns the agentview and the instruction into one embedding each. A context encoder fuses these inputs, and a flow-matching diffusion transformer attends to them to denoise an action chunk. The dashed decoder is used only to train the wrist encoder.}
\label{fig:arch}
\end{figure*}

The data separates where objects are from what the task is; the policy's input has to keep them separate. \ragevla{} therefore does not ask the action model to recover the layout from image tokens. Perception is a separate, supervised stage that outputs the layout explicitly, and the action model is conditioned on that output. Fig.~\ref{fig:arch} gives an overview. The policy sees LIBERO's standard agentview and wrist cameras, the robot's proprioceptive state and the instruction, and it outputs a chunk of 20 future actions. The system has two parts. The \emph{perception engine} (174M parameters) consists of DINOv3, shared by both cameras and counted once (86M), the two adapters (4M) and TinyCLIP (84M); only the adapters are trained. The \emph{action engine} is the context encoder and diffusion transformer, 35M parameters in \ragevla{} s1 and 100M in \ragevla{} n1, trained from scratch. The complete inference stack therefore has 0.21 and 0.27 billion parameters. Appendix~\ref{app:impl} gives layer-level details, losses and hyperparameters.

\subsection{Why DINOv3}
Both views are encoded by a frozen DINOv3 ViT-B/16~\cite{simeoni2025dinov3} at the cameras' native resolution. DINOv3 is trained with Gram anchoring, which keeps its dense patch features well localized over long self-supervised training. Its patch features therefore carry local geometry (edges, surface orientation, relative depth and part boundaries) as well as object identity, which is what a light head needs in order to locate objects in 3-D and to read depth. We combine features from several depths of the network: earlier layers are sharper in position, the last is stronger in identity. The backbone is never fine-tuned, so the adapters cannot overfit it to LIBERO's textures.

\subsection{Agentview object head}
A small convolutional head with skip connections, in the style of CenterNet~\cite{zhou2019centernet}, predicts for every cell of the feature grid a class heatmap together with the object's sub-cell offset, yaw and height. Heatmap peaks become detections. Each peak defines a camera ray, and intersecting the ray with the horizontal plane at the predicted height gives the object's 3-D position. The policy receives the result as a list of objects, each with its class, global position, yaw and confidence. Every label comes from the simulator.

\subsection{Wrist encoder}
The wrist image is encoded by the same frozen DINOv3. A small network compresses its features into a few tokens. To make this code useful for grasping, a decoder that is used only during training must reconstruct two dense maps from it: the class of what is visible in each cell, including surfaces, the gripper fingers and any held object, and its depth. The code must therefore keep what is under and around the gripper and how far away it is.

\subsection{Perception accuracy}
Table~\ref{tab:perc} reports both adapters on held-out frames, including frames from the held-out test set, whose layouts were never seen in training. The object head finds more than 99\% of objects with a median position error below 3~mm, so the policy receives global object positions that are accurate within a small tolerance wherever the objects are placed. The wrist code keeps the class of almost every cell under the gripper and its depth.

\begin{table}[t]
\centering
\caption{Perception accuracy on held-out frames (validation split and held-out test set).}
\label{tab:perc}
\footnotesize\setlength{\tabcolsep}{4pt}
\begin{tabular}{lcc}
\toprule
 & Validation & Held-out test \\
\midrule
Objects found & 99.3\% & 99.5\% \\
Object position error, median & 0.26 cm & 0.27 cm \\
Wrist cell class accuracy & \multicolumn{2}{c}{99.8\%} \\
Wrist object cell accuracy & \multicolumn{2}{c}{99.4\%} \\
Wrist depth error, median & \multicolumn{2}{c}{0.31 cm} \\
\bottomrule
\end{tabular}
\end{table}

\subsection{Vision-language conditioning with TinyCLIP}
\label{sec:clip}
Beyond the two perception adapters, all visual-semantic and linguistic information that the policy receives comes from TinyCLIP~\cite{wu2023tinyclip}, a distilled CLIP~\cite{radford2021clip}. Its image tower embeds the agentview into one vector and its text tower embeds the instruction into another. Because the text tower is frozen, all instructions and paraphrases are embedded once and stored; an unseen instruction is embedded at inference.

The contrast with VLM-based VLAs is one of scale. Their actions are computed from hundreds of high-dimensional tokens per image, produced by a backbone with billions of parameters. \ragevla{} receives two 512-dimensional vectors from an 84M-parameter encoder. The larger models bring broad pre-trained knowledge that \ragevla{} lacks. We do not claim that the small encoder is better. Our claim is narrower: on LIBERO-PRO, pre-trained knowledge did not prevent memorization, while a small aligned embedding combined with explicit perception and varied data did not prevent generalization.

The image embedding is partly redundant with the object list and the wrist code, and we keep it deliberately. During training each input group (objects, wrist, image, text) is dropped independently with a small probability, which discourages reliance on any single channel. Sec.~\ref{sec:probe} measures how the trained policy uses each of them.

\subsection{Flow-matching diffusion transformer}
Each input becomes one or more tokens: one per detected object, eight for the wrist code, and one each for the image embedding, the text embedding and the robot state. A small transformer encoder fuses them into a context. The denoiser is a diffusion transformer~\cite{peebles2023dit} over the action tokens of the chunk. Each block applies self-attention, cross-attention to the context and a feed-forward layer, and is modulated by the flow time through the shared conditioning scheme of PixArt-$\alpha$~\cite{chen2024pixart}, which saves parameters that we spend on depth. We train two sizes, \ragevla{} s1 with 35M parameters and \ragevla{} n1 with 100M (Appendix~\ref{app:impl}).

The model is trained from scratch with the flow-matching objective~\cite{lipman2023flow,liu2023rectified}: it learns the velocity that carries noise to the action chunk, integrates it in a few steps at inference, executes the first half of the chunk and re-plans. On the held-out test set the action error equals the validation error (Appendix~\ref{app:impl}), so the policy is not fitted to the trajectories it was trained on.

\section{Experiments}
\label{sec:exp}
\subsection{Protocol}
We follow LIBERO-PRO's single-perturbation evaluation, in which one perturbation is active at a time and every cell of the table is 10 tasks with 50 initial states each, 12{,}000 episodes in total. For Obj, Pos, Sem and Task we use the released task and initial-state files. No Env files are released, so we generate them with LIBERO-PRO's own perturbation code. The Pos column is the benchmark's swap test, as on the leaderboard, and the policy always receives the instruction stored in the evaluated task file. Appendix~\ref{app:protocol} explains why both points matter. Episode limits and success checks are the official ones, and colliding initial states are not filtered. Baseline numbers in the table are copied unchanged from the LIBERO-PRO leaderboard~\cite{liberopro_leaderboard}; the only baseline we ran ourselves is $\pi_{0.5}$ on LIBERO-Object (Sec.~\ref{sec:probe}). We report one training seed. Both \ragevla{} s1 and \ragevla{} n1 were evaluated on all 12{,}000 episodes with zero errors; charts show \ragevla{} s1. \prelimnote

\begin{table*}[t]
\centering
\caption{LIBERO-PRO success rates with one perturbation active at a time. Prior VLAs are taken from the LIBERO-PRO leaderboard, and a dash means not reported. The best value per column is in bold. \prelimnote}
\label{tab:main}
\setlength{\tabcolsep}{2.2pt}\scriptsize
\begin{tabular}{l|ccccc|ccccc|ccccc|ccccc|c}
\toprule
 & \multicolumn{5}{c|}{LIBERO-Goal} & \multicolumn{5}{c|}{LIBERO-Spatial} & \multicolumn{5}{c|}{LIBERO-10} & \multicolumn{5}{c|}{LIBERO-Object} & \\
Model & Obj & Pos & Sem & Task & Env & Obj & Pos & Sem & Task & Env & Obj & Pos & Sem & Task & Env & Obj & Pos & Sem & Task & Env & Total \\
\midrule
OpenVLA & 0.96 & 0.00 & \textbf{0.98} & 0.00 & \textbf{0.98} & \textbf{0.97} & 0.00 & \textbf{0.97} & 0.00 & 0.89 & 0.81 & 0.00 & \textbf{0.96} & 0.00 & 0.85 & \textbf{0.98} & 0.00 & 0.98 & 0.00 & 0.00 & 0.52 \\
$\pi_0$ & 0.94 & 0.00 & 0.93 & 0.00 & 0.39 & 0.95 & 0.00 & \textbf{0.97} & 0.00 & 0.60 & 0.79 & 0.00 & 0.82 & 0.00 & 0.27 & 0.94 & 0.00 & 0.90 & 0.00 & 0.29 & 0.44 \\
$\pi_{0.5}$ & \textbf{0.97} & 0.38 & 0.97 & 0.00 & 0.46 & \textbf{0.97} & 0.20 & \textbf{0.97} & 0.01 & 0.46 & \textbf{0.92} & 0.08 & 0.93 & 0.01 & 0.46 & \textbf{0.98} & 0.17 & 0.96 & 0.01 & 0.73 & 0.53 \\
MolmoAct & 0.68 & 0.00 & 0.85 & 0.00 & -- & 0.90 & 0.00 & 0.88 & 0.00 & -- & 0.54 & 0.00 & 0.74 & 0.06 & -- & 0.92 & 0.06 & 0.96 & 0.00 & -- & 0.41 \\
NORA & 0.58 & 0.00 & 0.88 & 0.00 & -- & 0.92 & 0.00 & 0.91 & 0.00 & -- & 0.46 & 0.00 & 0.74 & 0.00 & -- & 0.86 & 0.00 & 0.92 & 0.00 & -- & 0.40 \\
X-VLA & 0.68 & 0.01 & \textbf{0.98} & 0.09 & -- & \textbf{0.97} & 0.00 & 0.96 & 0.00 & -- & 0.62 & 0.00 & 0.95 & 0.10 & -- & 0.89 & 0.02 & 0.98 & 0.08 & -- & 0.46 \\
\midrule
\textbf{RageVLA s1 (ours)} & 0.72 & 0.36 & 0.95 & \textbf{0.20} & 0.84 & 0.92 & 0.53 & 0.90 & 0.51 & 0.86 & 0.45 & \textbf{0.51} & 0.88 & \textbf{0.42} & 0.86 & 0.78 & 0.94 & \textbf{0.99} & \textbf{0.96} & \textbf{0.87} & 0.72 \\
\textbf{RageVLA n1 (ours)} & 0.79 & \textbf{0.38} & 0.95 & 0.17 & 0.85 & 0.94 & \textbf{0.62} & 0.84 & \textbf{0.54} & \textbf{0.94} & 0.48 & 0.43 & 0.91 & 0.38 & \textbf{0.88} & 0.68 & \textbf{0.96} & 0.99 & \textbf{0.96} & 0.87 & \textbf{0.73} \\
\bottomrule
\end{tabular}
\end{table*}

\subsection{Main results}
\begin{figure*}[t]
\centering
\includegraphics[width=\textwidth]{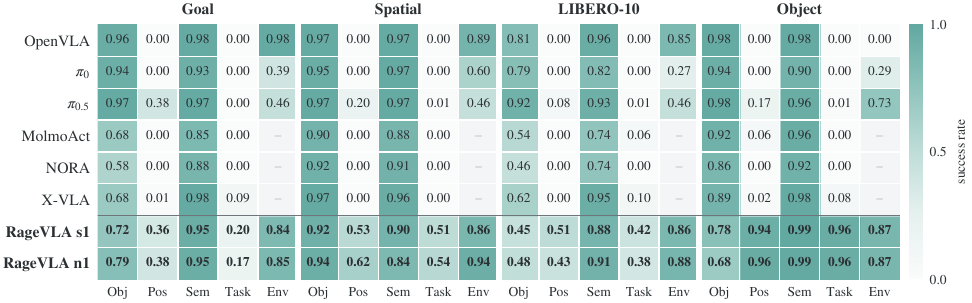}
\caption{LIBERO-PRO success per suite and perturbation. Rows above the line are published VLAs from the leaderboard, and a dash means not reported. \prelimnote}
\label{fig:heatmap}
\end{figure*}

Table~\ref{tab:main} and Fig.~\ref{fig:heatmap} give the full table. \ragevla{} s1 has the highest total, 0.72 against 0.53 for $\pi_{0.5}$. Its advantage is concentrated where prior models fail. It is ahead of every prior model in every Task cell, on Pos in three of four suites (0.53, 0.51 and 0.94 on Spatial, LIBERO-10 and Object, against at most 0.20 for any prior model), and on Env in two suites. On Obj and Sem, where most models already score highly, it is comparable but not better. The leaderboard's Task column depends on which instruction the evaluated policy was given (Appendix~\ref{app:protocol}). Against an independent re-evaluation of $\pi_{0.5}$ that uses the same instruction source as we do, the Task mean is 0.52 against 0.26 rather than against 0.01. Its weakest cell is Obj on LIBERO-10, which is consistent with the absence of appearance augmentation: appearance is the factor that our data leaves coupled. \ragevla{} n1 reaches a total of 0.73 with the same profile: it is stronger on Spatial Pos (0.62) and on Spatial and LIBERO-10 Env, and weaker on Object Obj. With an action engine three times larger, it is only 0.01 ahead of s1, which suggests that the inputs and the data, not the size of the action model, limit the scores.

\subsection{Reading the table without memorization}
Table~\ref{tab:memfree} splits LIBERO-PRO's columns. On Obj and Sem every compared model scores highly, and \ragevla{} s1 is in the same range. On Pos and Task the six prior VLAs average between 0.00 and 0.11, and \ragevla{} s1 averages 0.55 with a far smaller model. Including Env, which only three prior models report, \ragevla{} s1 reaches 0.65 against at most 0.25. \ragevla{} n1 has 0.56 and 0.66.

\begin{table}[t]
\centering
\caption{Splitting the LIBERO-PRO total. Obj and Sem leave layout and task unchanged, so models that replay training trajectories still score highly on them. The columns that move objects or change the task separate the models. \prelimnote}
\label{tab:memfree}
\scriptsize\setlength{\tabcolsep}{2.5pt}
\begin{tabular}{lccccc}
\toprule
Model & Params & Total & Obj+Sem & Pos+Task & Pos+Task+Env \\
\midrule
OpenVLA & 7.50B & 0.52 & 0.95 & 0.00 & 0.23 \\
$\pi_0$ & 3.30B & 0.44 & 0.90 & 0.00 & 0.13 \\
$\pi_{0.5}$ & 3.30B & 0.53 & 0.96 & 0.11 & 0.25 \\
MolmoAct & 8.10B & 0.41 & 0.81 & 0.01 & -- \\
NORA & 3.80B & 0.40 & 0.78 & 0.00 & -- \\
X-VLA & 0.90B & 0.46 & 0.88 & 0.04 & -- \\
\midrule
\textbf{RageVLA s1 (ours)} & 0.21B & 0.72 & 0.83 & 0.55 & 0.65 \\
\textbf{RageVLA n1 (ours)} & 0.27B & 0.73 & 0.82 & 0.56 & 0.66 \\
\bottomrule
\end{tabular}
\end{table}

\subsection{Does the policy act on what it perceives?}
\label{sec:probe}
The LIBERO-PRO table shows that \ragevla{} succeeds under perturbation. It does not show how: a policy could still replay trajectories that happen to cover the test layouts. We test this directly on LIBERO-Object with \ragevla{} s1. Every number below is from 10 tasks with 20 initial states each, 200 episodes, under the protocol of the main table (95\% binomial half-width at most 0.07).

\begin{figure}[t]
\centering
\includegraphics[width=\columnwidth]{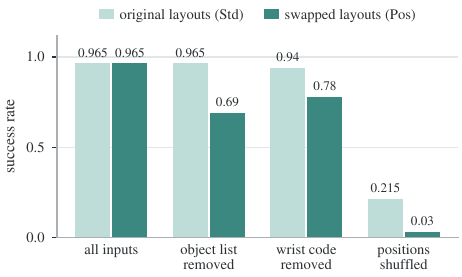}
\caption{\ragevla{} s1 on LIBERO-Object with one input changed at inference, 200 episodes per bar. The camera images are never changed. Shuffling gives each detected object the position of another one.}
\label{fig:ablation}
\end{figure}

\parhead{Changing one input at inference} We alter one input of the action model and leave everything else untouched, including the camera images (Fig.~\ref{fig:ablation}). \emph{Removing} the object list or the wrist code applies the mask that input dropout applies during training. \emph{Shuffling} gives every detected object the position and yaw of the next detected object and keeps its class, so the policy receives a complete but wrong layout.

Shuffling is the decisive case. Success falls from 0.965 to 0.215 in LIBERO's original layouts and from 0.965 to 0.03 in the swapped ones. The original layouts are where a replayed trajectory would work best, and the images there are unchanged, yet the policy follows the positions it is given and fails. Its actions are computed from the object list, not recalled.

Removing an input shows the redundancy that the design intends. In the original layouts the policy loses nothing without the object list (0.965) and little without the wrist code (0.94). In the swapped layouts it falls to 0.69 and 0.78: both inputs carry the robustness to moved objects, and neither is a single point of failure, because the image embedding and the remaining input still locate the target in most episodes.

\parhead{Graded displacement} LIBERO-PRO also ships ten LIBERO-Object suites in which the object to be picked is translated by 7 to 35~cm along one table axis (Appendix~\ref{app:protocol}). They measure robustness as a function of distance and are not part of the leaderboard. We run \ragevla{} s1 on them and, as a reference, the official $\pi_{0.5}$ LIBERO checkpoint of openpi~\cite{openpi2025}, through the same harness, episodes and success checks, with observations built as in openpi's own LIBERO evaluation. In this harness $\pi_{0.5}$ solves all 200 unperturbed episodes and 0.195 of the swap episodes, close to its leaderboard value of 0.17, so the harness reproduces its published behaviour.

\begin{figure}[t]
\centering
\includegraphics[width=\columnwidth]{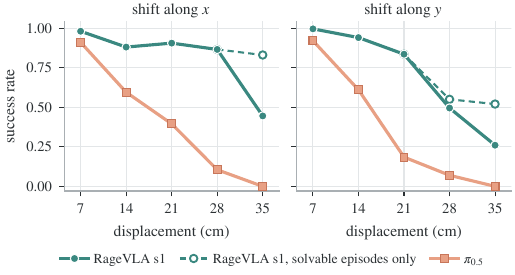}
\caption{Success on LIBERO-Object when the object to be picked is displaced (LIBERO-PRO's graded suites), 200 episodes per point, both models in the same harness. Solid: all episodes. Dashed: without the episodes that no policy can solve, because the object starts out of reach, the basket is moved off the table or the file cannot be loaded.}
\label{fig:shift}
\end{figure}

Fig.~\ref{fig:shift} shows the result. $\pi_{0.5}$ loses success steadily with distance: 0.60 and 0.61 at 14~cm along $x$ and $y$, 0.11 and 0.07 at 28~cm, and none at 35~cm. \ragevla{} s1 stays at 0.865 or above up to 28~cm along $x$, and at 0.835 or above up to 21~cm along $y$.

At the largest displacements part of the drop belongs to the benchmark files and not to the policy. At 35~cm along $x$ the object starts beyond the reach of the arm in five of the ten tasks. Along $y$ the files move the basket off the table in one task at 28~cm and in four at 35~cm, and one file at 35~cm cannot be loaded. These episodes cannot be solved by any policy, and we count them as failures in the solid curves. On the remaining episodes \ragevla{} s1 succeeds in 0.83 at 35~cm along $x$, and in 0.55 and 0.52 at 28 and 35~cm along $y$, while $\pi_{0.5}$ remains at 0.00 at 35~cm (Table~\ref{tab:shiftres}). Large displacements along $y$ are the weak case of \ragevla{} s1.

\parhead{What this shows} The comparison is between two complete systems. $\pi_{0.5}$ was trained on LIBERO's original demonstrations and \ragevla{} s1 on our augmented data, so the gap is the effect of decoupled data and decoupled inputs together. What the two tests establish is the mechanism: the behaviour of \ragevla{} s1 is determined by the object positions it perceives, in the original layouts as well as in perturbed ones, and it holds as the displacement grows.

\subsection{Standard LIBERO}
On the unperturbed tasks of the same runs both \ragevla{} s1 and n1 succeed 0.96 of the time (Appendix, Table~\ref{tab:std}), slightly below the best reported LIBERO numbers. We read this as consistent with the training objective: the policy never sees a demonstration replayed in its original layout often enough to memorize it, since almost every training trajectory is laid out anew. The few points it gives up on standard LIBERO are the same property that lets it keep most of its success when objects move, the environment changes or the task changes.

\subsection{Comparison with an LLM coding agent}
\label{sec:aspire}
\begin{figure}[t]
\centering
\includegraphics[width=\columnwidth]{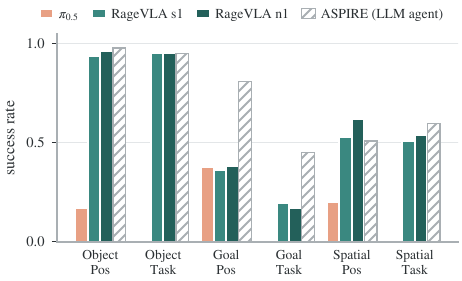}
\caption{Pos and Task on the three suites reported by ASPIRE~\cite{aspire2026}. ASPIRE synthesizes a program per task with a frontier language model, and the other two are learned policies. \prelimnote}
\label{fig:aspire}
\end{figure}

ASPIRE~\cite{aspire2026} reports high success on LIBERO-PRO's Pos and Task perturbations (Fig.~\ref{fig:aspire}), and close to full success with its strongest language model. The two approaches answer different questions. ASPIRE trains no network weights. For each task, a frontier language model writes a program against perception, geometry and motion-planning interfaces, debugs it on a set of practice layouts, and distills validated repairs into a reusable skill library. This synthesis has a per-task cost in agent time and API usage. In the authors' model comparison on the ten LIBERO-PRO Goal tasks under the position perturbation, run without their evolutionary search, the two most successful models need roughly an hour of agent time and \$12 to \$18 per task, whereas \ragevla{} produces an action chunk in milliseconds. \ragevla{} is a learned closed-loop policy that runs a small network at every control step and needs no per-task program synthesis before deployment. The protocols also differ, since ASPIRE evaluates on its own seeded layouts and uses depth and segmentation inputs, so the comparison is indicative rather than controlled. We therefore compare against VLA policies and report ASPIRE as a reference for what program synthesis with a frontier model can reach.

\section{Limitations}
\ragevla{} is trained on 130 simulated tasks, and all results are in simulation. It does not understand goals outside its training distribution, and we do not claim generality beyond LIBERO. The object head detects LIBERO's object and fixture classes only.

The perception adapters are supervised with simulator labels and use the known camera calibration. On a real robot both would need a new source. No color or texture augmentation was used, which shows in the Obj column on LIBERO-10.

The augmented data and the perception interface were designed together and are evaluated together; we do not measure the contribution of either one alone. The tests of Sec.~\ref{sec:probe} change inputs at inference, cover one suite with 200 episodes per number, and compare against a $\pi_{0.5}$ checkpoint trained on the original demonstrations. Displacements of 28~cm and more along the $y$ axis still roughly halve the success of \ragevla{} s1.

On the evaluation side, baseline numbers in the main table come from the LIBERO-PRO leaderboard rather than from our own runs, and our Env column uses files generated with LIBERO-PRO's released code (Appendix~\ref{app:protocol}). Results are from one training seed.

\section{Conclusion}
On LIBERO-PRO, what separates policies is not their parameter count but whether what they do is decoupled from where things were during training. \ragevla{} decouples twice: its training data varies layout, environment, surface and wording by general rules, and its inputs present object positions, wrist geometry and the instruction as separate signals. A 0.21-billion-parameter policy built this way, with an action engine of 35M parameters trained from scratch, achieves the best LIBERO-PRO result among VLAs and is the only one with substantial success under both position and task perturbations. Its actions follow the object positions it perceives, and it keeps most of its success when the target is displaced by up to 28~cm along one axis, where $\pi_{0.5}$ has lost almost all of it. The next steps are appearance augmentation, training data with far more than 130 tasks, and a real robot.

\section*{Acknowledgment}
This work uses DINO Materials (DINOv3~\cite{simeoni2025dinov3}) provided by Meta under the DINOv3 License.

\appendices
\section{Implementation details}
\label{app:impl}
\parhead{Augmentation} About 70\% of attempts use the uniform layout and 30\% the smaller displacements. Per round each task receives about 400 successful trajectories: 240 in its own environment, 100 in other environments and 60 on other surfaces; a task without a valid transplant puts that quota into its own environment. Fixture yaw is uniform in $\pm180^\circ$, object yaw follows each object's rotational symmetry. Each generator attempt is re-simulated and at most three attempts per requested success are made. Trajectories with no sample at stride 8 are dropped. Six tasks have paraphrases that change the meaning, which we kept.

\parhead{Agentview object head} The image is rendered at $256^2$ and encoded by DINOv3 ViT-B/16 into a $16\times16$ grid; features of layers 4, 8 and 12 enter a convolutional head with skip connections (1.7M parameters) whose $1\times1$ outputs are a heatmap over the object and fixture classes plus the gripper, a 2-D offset, the yaw as $(\sin k\psi,\cos k\psi)$ with $k$ set by the object's symmetry, and the height. The loss is
\begin{equation}
\mathcal{L}_{\text{agent}} = \mathcal{L}_{\text{focal}} + \mathcal{L}_{\text{off}} + \mathcal{L}_{\text{yaw}} + 5\,\mathcal{L}_{z},
\end{equation}
with the CenterNet focal loss~\cite{zhou2019centernet,lin2017focal} and L1 losses at positive cells. Both adapters are trained on 432{,}592 frames, six phased frames per training trajectory, with batch 256.

\parhead{Wrist encoder} The native $256^2$ wrist image is encoded by the same DINOv3. The 768-dimensional features are projected to 256 dimensions and compressed into eight 32-dimensional tokens (2.6M parameters). The training-only decoder (2.1M parameters) predicts per cell a class and a log-depth and refines them into a $64\times64$ log-depth map. The loss is
\begin{equation}
\begin{split}
\mathcal{L}_{\text{wrist}} = {}& \mathrm{CE}_w(\hat c, c) + 2\,\lVert \log\hat d_{16} - \log d_{16}\rVert_1 \\
&+ 2\,\lVert \log\hat d_{64} - \log d_{64}\rVert_1 ,
\end{split}
\end{equation}
where the class weights are the square root of inverse class frequency, clipped to $[0.2, 8]$.

\parhead{TinyCLIP} We use TinyCLIP-ViT-40M/32 with a 39.4M image tower and a 44.3M text tower. For comparison, the hidden states that VLM-based policies condition on have 2{,}048 to 4{,}096 dimensions per token.

\parhead{Flow matching} With a noise sample $x_0$, a target action chunk $a$ and a flow time $\tau \in (0,1)$, the model sees $x_\tau = (1-\tau)\,x_0 + \tau a$ and predicts the velocity $a - x_0$ under a squared error. At inference it integrates the velocity in ten Euler steps.

\parhead{Action model} Table~\ref{tab:hyper} lists the hyperparameters. Objects are embedded as a class embedding plus an MLP of Fourier features of the position, concatenated with yaw and score. Absent objects and dropped inputs are masked in the context encoder. The proprioceptive state is the end-effector position relative to the base, its axis-angle orientation and the two finger positions. On the held-out test set, \ragevla{} s1 has a sampled-chunk position error of 0.112 against 0.110 on validation.

\begin{table}[h]
\centering
\caption{Action model and training hyperparameters.}
\label{tab:hyper}
\footnotesize\setlength{\tabcolsep}{3pt}
\begin{tabular}{lll}
\toprule
 & \ragevla{} s1 & \ragevla{} n1 \\
\midrule
Width, heads & 384, 6 & 512, 8 \\
Denoiser blocks & 12 & 20 \\
Parameters & 35M & 100M \\
Peak learning rate & $6\cdot10^{-4}$ & $3\cdot10^{-4}$ \\
Warm-up steps & 500 & 2{,}000 \\
Epochs & 23 & 18 \\
\midrule
Action chunk & \multicolumn{2}{l}{20 actions of 7 dimensions} \\
Time conditioning & \multicolumn{2}{l}{adaLN-single~\cite{chen2024pixart}} \\
Flow time & \multicolumn{2}{l}{$\tau \sim \mathrm{sigmoid}(\mathcal{N}(0,1))$} \\
Optimizer & \multicolumn{2}{l}{AdamW, cosine decay} \\
EMA & \multicolumn{2}{l}{0.999} \\
Input dropout & \multicolumn{2}{l}{per input group} \\
Sampling & \multicolumn{2}{l}{10 Euler steps, 10 of 20 actions executed} \\
\bottomrule
\end{tabular}
\end{table}

\section{Additional results}
\label{app:results}
\begin{table}[h]
\centering
\caption{\ragevla{} on the unperturbed tasks, from the same runs. \prelimnote}
\label{tab:std}
\footnotesize
\begin{tabular}{lccccc}
\toprule
& Goal & Spatial & LIBERO-10 & Object & Mean \\
\midrule
s1 & 0.96 & 0.96 & 0.95 & 0.98 & 0.96 \\
n1 & 0.97 & 0.95 & 0.92 & 0.98 & 0.96 \\
\bottomrule
\end{tabular}
\end{table}

\begin{figure*}[t]
\begin{minipage}[t]{\columnwidth}
\centering
\includegraphics[width=\columnwidth]{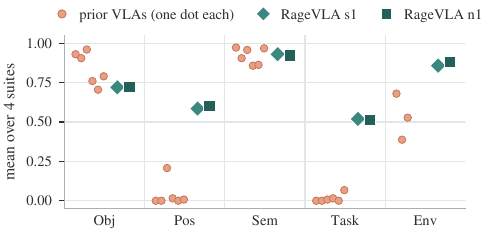}
\caption{Success per perturbation type, averaged over the four suites. Models that do not report Env are omitted there. \prelimnote}
\label{fig:perdim}
\end{minipage}\hfill
\begin{minipage}[t]{\columnwidth}
\centering
\includegraphics[width=\columnwidth]{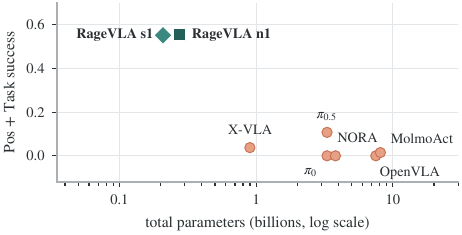}
\caption{Success on Pos and Task against total parameter count, including vision encoders. \prelimnote}
\label{fig:scale}
\end{minipage}
\end{figure*}

\begin{figure}[t]
\centering
\includegraphics[width=\columnwidth]{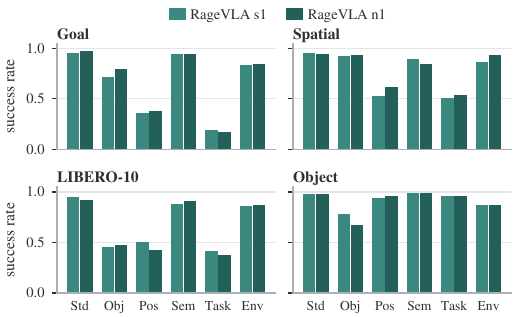}
\caption{\ragevla{} per suite and column, including the unperturbed tasks (Std), for both model sizes. \prelimnote}
\label{fig:suites}
\end{figure}

Fig.~\ref{fig:perdim} shows the mean of every model per perturbation type, Fig.~\ref{fig:scale} plots success on Pos and Task against model size, and Fig.~\ref{fig:suites} gives the results of both models per suite. The policy samples fresh noise at every re-plan, so each cell is an estimate from 500 stochastic episodes with a 95\% binomial half-width of at most 0.045.

\parhead{Graded shift results} Table~\ref{tab:shiftres} gives the numbers behind Fig.~\ref{fig:shift}. $\pi_{0.5}$ is the official \texttt{pi05\_libero} checkpoint~\cite{openpi2025}; its observations (images padded and resized to 224 pixels, end-effector position, axis-angle orientation and gripper state) and its re-planning interval of five steps follow openpi's LIBERO evaluation. Both models use the first 20 released initial states of every task, the instruction stored in the task file, and LIBERO's step limit and success check for the suite. An episode is counted as infeasible, independently of the policy, if the object to be picked starts more than 0.855~m from the robot base, if the file places the basket off the table, or if the environment cannot be built from the file; the last case is one task at 35~cm along $y$, whose placement region is empty.

\begin{table}[h]
\centering
\caption{LIBERO-Object under LIBERO-PRO's graded object shift, 200 episodes per level. ``Feasible'' excludes the episodes that no policy can solve; their number is in the last column.}
\label{tab:shiftres}
\scriptsize\setlength{\tabcolsep}{3pt}
\begin{tabular}{llccccc}
\toprule
 & & \multicolumn{2}{c}{\ragevla{} s1} & \multicolumn{2}{c}{$\pi_{0.5}$} & \\
Axis & Shift & All & Feasible & All & Feasible & Infeasible \\
\midrule
$x$ & 7 cm & 0.980 & 0.980 & 0.910 & 0.910 & 0 \\
$x$ & 14 cm & 0.880 & 0.880 & 0.595 & 0.595 & 0 \\
$x$ & 21 cm & 0.905 & 0.905 & 0.395 & 0.395 & 0 \\
$x$ & 28 cm & 0.865 & 0.865 & 0.105 & 0.105 & 0 \\
$x$ & 35 cm & 0.445 & 0.830 & 0.000 & 0.000 & 100 \\
\midrule
$y$ & 7 cm & 0.995 & 0.995 & 0.920 & 0.920 & 0 \\
$y$ & 14 cm & 0.940 & 0.940 & 0.610 & 0.610 & 0 \\
$y$ & 21 cm & 0.835 & 0.835 & 0.185 & 0.185 & 0 \\
$y$ & 28 cm & 0.495 & 0.550 & 0.070 & 0.078 & 20 \\
$y$ & 35 cm & 0.260 & 0.520 & 0.000 & 0.000 & 100 \\
\midrule
\multicolumn{2}{l}{Unperturbed (Std)} & 0.965 & & 1.000 & & \\
\multicolumn{2}{l}{Swap (Pos)} & 0.965 & & 0.195 & & \\
\bottomrule
\end{tabular}
\end{table}

On the swap suite \ragevla{} s1 scores 0.965 here and 0.94 in Table~\ref{tab:main}; the checkpoint is the same, and the difference is 20 against 50 initial states per task.

\section{Notes on the LIBERO-PRO protocol}
\label{app:protocol}
Two details of the benchmark's released code affect how its columns should be read. We state what we evaluated and what we could establish about the published entries.

\parhead{Two position tests} LIBERO-PRO contains two different position perturbations, and only one of them is in the leaderboard. (i)~\emph{Swap}: the suites \texttt{libero\_\{goal, spatial, 10, object\}\_swap}, selected with \texttt{use\_swap}, exchange the initial region of a task object with that of one object from a fixed candidate list (\texttt{ood\_spatial\_relation.yaml}). BDDL and initial-state files are released for all 40 tasks and the perturbation has no magnitude parameter. The reference evaluation flow maps the Pos setting to these suites; this is the Pos column of Table~\ref{tab:main}, for prior models and for \ragevla{} s1. (ii)~\emph{Graded shift}: for LIBERO-Object only, the repository ships ten further suites (\texttt{libero\_object\_temp\_x0.1}--\texttt{x0.5} and \texttt{y0.1}--\texttt{y0.5}) in which the initial region of the object to be picked is translated along one table axis. In the released files one 0.1 step is 7~cm, so the five levels are 7, 14, 21, 28 and 35~cm (Table~\ref{tab:shift}); objects that would overlap the new region are moved off the table, and at the two largest $y$ levels this includes the basket in one and four of the ten tasks. These suites underlie the displacement-sensitivity curve of~\cite{zhou2025liberopro} (Fig.~4 of its v2, in undefined ``units'': OpenVLA and $\pi_0$ fail beyond 0.2, $\pi_{0.5}$ beyond 0.4). No levels exist for the other three suites and no leaderboard entry reports them. A later dataset release adds an initial position-and-yaw case for all 40 tasks as part of a seven-case robustness extension; it is not in the leaderboard either. All Pos numbers in this paper refer to (i); the suites of (ii) are evaluated separately in Sec.~\ref{sec:probe}.

\begin{table}[h]
\centering
\caption{The two position tests in LIBERO-PRO's release.}
\label{tab:shift}
\scriptsize\setlength{\tabcolsep}{3pt}
\begin{tabular}{lll}
\toprule
 & Swap (leaderboard Pos) & Graded shift \\
\midrule
Suites & all four (40 tasks) & LIBERO-Object (10 tasks) \\
What moves & 2--4 objects exchange regions & picked object only \\
Magnitude & fixed by the object pair & 7, 14, 21, 28, 35~cm; $x$ or $y$ \\
Reported in & leaderboard, Table~\ref{tab:main} & Fig.~4 of~\cite{zhou2025liberopro} \\
Used here & Table~\ref{tab:main} & Sec.~\ref{sec:probe} \\
\bottomrule
\end{tabular}
\end{table}

\parhead{Environment files} LIBERO-PRO's Env perturbation (\texttt{use\_environment}, \texttt{ood\_environment.yaml}) rewrites the scene of a task so that it takes place on the living-room table; five LIBERO-10 tasks are already there and stay unchanged. The release contains no task or initial-state files for this perturbation. We produced them with the release's own perturbation code, with only this flag set, and drew 50 initial states per task with the release's initial-state procedure, without filtering.

\parhead{Which instruction the policy receives} The Sem and Task perturbations rewrite the \texttt{(:language)} field of the BDDL (Task also rewrites the goal). The benchmark's task loader derives the instruction from the BDDL \emph{file name}, which the perturbations leave unchanged, and the usual runners pass that string to the policy~\cite{liberopro_issue14,liberopro_issue27}. With such a runner the policy is given the original instruction in both columns: Sem then equals the unperturbed suite, and Task checks a new goal while stating the old one. The maintainers recommend reading the instruction from the environment instead~\cite{liberopro_issue14}, and that is what we do, so our Sem column tests the paraphrases and our Task column states the new goal. 

We cannot determine which instruction source produced each leaderboard entry. An independent re-evaluation of the public $\pi_{0.5}$ LIBERO checkpoint~\cite{liberopro_issue27} comes close to the leaderboard total with file-name instructions (0.54 against 0.53) and reports that, with BDDL instructions, Sem is unchanged (0.96) while Task rises from 0.04 to 0.26 (Table~\ref{tab:instr}). Against that figure \ragevla{} s1's Task advantage is 0.52 to 0.26 rather than 0.52 to 0.01; it is slightly lower on Goal and Spatial. The same source reports 0.27 for $\pi_{0.5}$ on Pos, averaged over suites (leaderboard: 0.21; \ragevla{} s1: 0.59). These figures come from an issue thread, not from a peer-reviewed source, and we did not re-run them; we include them because they bear directly on how the Task column is read.

\begin{table}[h]
\centering
\caption{Task perturbation by instruction source. $\pi_{0.5}$ rows: leaderboard~\cite{liberopro_leaderboard} and an independent re-evaluation with the BDDL instruction~\cite{liberopro_issue27} (100 episodes per task). }
\label{tab:instr}
\scriptsize\setlength{\tabcolsep}{3pt}
\begin{tabular}{lccccc}
\toprule
 & Goal & Spatial & LIBERO-10 & Object & Mean \\
\midrule
$\pi_{0.5}$, leaderboard & 0.00 & 0.01 & 0.01 & 0.01 & 0.01 \\
$\pi_{0.5}$, BDDL instruction~\cite{liberopro_issue27} & 0.22 & 0.53 & 0.18 & 0.10 & 0.26 \\
\ragevla{} s1, BDDL instruction & 0.20 & 0.51 & 0.42 & 0.96 & 0.52 \\
\bottomrule
\end{tabular}
\end{table}

\bibliographystyle{IEEEtran}
\bibliography{refs}

\begin{thebibliography}{10}
\providecommand{\url}[1]{#1}
\csname url@samestyle\endcsname
\providecommand{\newblock}{\relax}
\providecommand{\bibinfo}[2]{#2}
\providecommand{\BIBentrySTDinterwordspacing}{\spaceskip=0pt\relax}
\providecommand{\BIBentryALTinterwordstretchfactor}{4}
\providecommand{\BIBentryALTinterwordspacing}{\spaceskip=\fontdimen2\font plus
\BIBentryALTinterwordstretchfactor\fontdimen3\font minus
  \fontdimen4\font\relax}
\providecommand{\BIBforeignlanguage}[2]{{%
\expandafter\ifx\csname l@#1\endcsname\relax
\typeout{** WARNING: IEEEtran.bst: No hyphenation pattern has been}%
\typeout{** loaded for the language `#1'. Using the pattern for}%
\typeout{** the default language instead.}%
\else
\language=\csname l@#1\endcsname
\fi
#2}}
\providecommand{\BIBdecl}{\relax}
\BIBdecl

\bibitem{liu2023libero}
B.~Liu, Y.~Zhu, C.~Gao, Y.~Feng, Q.~Liu, Y.~Zhu, and P.~Stone, ``{LIBERO}:
  Benchmarking knowledge transfer for lifelong robot learning,'' in
  \emph{Advances in Neural Information Processing Systems (Datasets and
  Benchmarks)}, 2023.

\bibitem{kim2025oft}
M.~J. Kim, C.~Finn, and P.~Liang, ``Fine-tuning vision-language-action models:
  Optimizing speed and success,'' \emph{arXiv preprint arXiv:2502.19645}, 2025.

\bibitem{openpi2025}
{Physical Intelligence}, ``openpi: {LIBERO} results for $\pi_{0.5}$,''
  \url{https://github.com/Physical-Intelligence/openpi/blob/main/examples/libero/README.md},
  2025.

\bibitem{zheng2025xvla}
J.~Zheng \emph{et~al.}, ``{X-VLA}: Soft-prompted transformer as scalable
  cross-embodiment vision-language-action model,'' \emph{arXiv preprint
  arXiv:2510.10274}, 2025.

\bibitem{zhou2025liberopro}
X.~Zhou, Y.~Xu, G.~Tie, Y.~Chen, G.~Zhang, D.~Chu, P.~Zhou, and L.~Sun,
  ``{LIBERO-PRO}: Towards robust and fair evaluation of vision-language-action
  models beyond memorization,'' \emph{arXiv preprint arXiv:2510.03827}, 2025.

\bibitem{kim2024openvla}
M.~J. Kim, K.~Pertsch, S.~Karamcheti, T.~Xiao, A.~Balakrishna, S.~Nair,
  R.~Rafailov, E.~Foster, G.~Lam, P.~Sanketi \emph{et~al.}, ``{OpenVLA}: An
  open-source vision-language-action model,'' \emph{arXiv preprint
  arXiv:2406.09246}, 2024.

\bibitem{black2024pi0}
K.~Black, N.~Brown, D.~Driess \emph{et~al.}, ``$\pi_0$: A
  vision-language-action flow model for general robot control,'' \emph{arXiv
  preprint arXiv:2410.24164}, 2024.

\bibitem{lee2025molmoact}
J.~Lee \emph{et~al.}, ``{MolmoAct}: Action reasoning models that can reason in
  space,'' \emph{arXiv preprint arXiv:2508.07917}, 2025.

\bibitem{hung2025nora}
C.-Y. Hung \emph{et~al.}, ``{NORA}: A small open-sourced generalist vision
  language action model for embodied tasks,'' \emph{arXiv preprint
  arXiv:2504.19854}, 2025.

\bibitem{pi2025pi05}
{Physical Intelligence}, K.~Black \emph{et~al.}, ``$\pi_{0.5}$: a
  vision-language-action model with open-world generalization,'' \emph{arXiv
  preprint arXiv:2504.16054}, 2025.

\bibitem{simeoni2025dinov3}
O.~Sim{\'e}oni, H.~V. Vo, M.~Seitzer, F.~Baldassarre, M.~Oquab \emph{et~al.},
  ``{DINOv3},'' \emph{arXiv preprint arXiv:2508.10104}, 2025.

\bibitem{wu2023tinyclip}
K.~Wu, H.~Peng, Z.~Zhou \emph{et~al.}, ``{TinyCLIP}: {CLIP} distillation via
  affinity mimicking and weight inheritance,'' in \emph{IEEE/CVF International
  Conference on Computer Vision}, 2023.

\bibitem{shukor2025smolvla}
M.~Shukor \emph{et~al.}, ``{SmolVLA}: A vision-language-action model for
  affordable and efficient robotics,'' \emph{arXiv preprint arXiv:2506.01844},
  2025.

\bibitem{octo2024}
D.~Ghosh, H.~R. Walke, K.~Pertsch \emph{et~al.}, ``Octo: An open-source
  generalist robot policy,'' in \emph{Robotics: Science and Systems}, 2024.

\bibitem{radford2021clip}
A.~Radford, J.~W. Kim, C.~Hallacy \emph{et~al.}, ``Learning transferable visual
  models from natural language supervision,'' in \emph{International Conference
  on Machine Learning}, 2021.

\bibitem{fei2025liberoplus}
S.~Fei \emph{et~al.}, ``{LIBERO-Plus}: In-depth robustness analysis of
  vision-language-action models,'' \emph{arXiv preprint arXiv:2510.13626},
  2025.

\bibitem{mandlekar2023mimicgen}
A.~Mandlekar, S.~Nasiriany, B.~Wen \emph{et~al.}, ``{MimicGen}: A data
  generation system for scalable robot learning using human demonstrations,''
  in \emph{Conference on Robot Learning}, 2023.

\bibitem{chi2023diffusionpolicy}
C.~Chi, S.~Feng, Y.~Du, Z.~Xu, E.~Cousineau, B.~Burchfiel, and S.~Song,
  ``Diffusion policy: Visuomotor policy learning via action diffusion,'' in
  \emph{Robotics: Science and Systems}, 2023.

\bibitem{peebles2023dit}
W.~Peebles and S.~Xie, ``Scalable diffusion models with transformers,'' in
  \emph{IEEE/CVF International Conference on Computer Vision}, 2023.

\bibitem{lipman2023flow}
Y.~Lipman, R.~T.~Q. Chen, H.~Ben-Hamu, M.~Nickel, and M.~Le, ``Flow matching
  for generative modeling,'' in \emph{International Conference on Learning
  Representations}, 2023.

\bibitem{liu2023rectified}
X.~Liu, C.~Gong, and Q.~Liu, ``Flow straight and fast: Learning to generate and
  transfer data with rectified flow,'' in \emph{International Conference on
  Learning Representations}, 2023.

\bibitem{chen2024pixart}
J.~Chen, J.~Yu, C.~Ge \emph{et~al.}, ``{PixArt}-$\alpha$: Fast training of
  diffusion transformer for photorealistic text-to-image synthesis,'' in
  \emph{International Conference on Learning Representations}, 2024.

\bibitem{zhou2019centernet}
X.~Zhou, D.~Wang, and P.~Kr{\"a}henb{\"u}hl, ``Objects as points,'' \emph{arXiv
  preprint arXiv:1904.07850}, 2019.

\bibitem{liang2023cap}
J.~Liang, W.~Huang, F.~Xia \emph{et~al.}, ``Code as policies: Language model
  programs for embodied control,'' in \emph{IEEE International Conference on
  Robotics and Automation}, 2023.

\bibitem{aspire2026}
R.~Lu, Y.~Wu, E.~Kou, H.~Lin, L.~Fu, W.~Xiao, A.~Mandlekar, Y.~Xu, G.~Shi,
  K.~Goldberg, A.~Chen, M.~Chowdhury, J.~Kautz, Y.~Zhu, L.~Fan, and G.~Wang,
  ``{ASPIRE}: Agentic {/Skills} discovery for robotics,'' \emph{arXiv preprint
  arXiv:2607.00272}, 2026, v2, Oct.~2026.

\bibitem{liberopro_leaderboard}
X.~Zhou, Y.~Xu, G.~Tie, Y.~Chen, G.~Zhang, D.~Chu, P.~Zhou, and L.~Sun,
  ``{LIBERO-PRO} model leaderboard,''
  \url{https://github.com/Zxy-MLlab/LIBERO-PRO}, 2025, accessed Nov.~2026.

\bibitem{lin2017focal}
T.-Y. Lin, P.~Goyal, R.~Girshick, K.~He, and P.~Doll{\'a}r, ``Focal loss for
  dense object detection,'' in \emph{IEEE International Conference on Computer
  Vision}, 2017.

\bibitem{liberopro_issue14}
{LIBERO-PRO community}, ``Is there a bug related to the `{TASK}' perturbation?
  ({LIBERO-PRO} issue \#14, with maintainer reply),''
  \url{https://github.com/Zxy-MLlab/LIBERO-PRO/issues/14}, 2025, accessed
  Nov.~2026.

\bibitem{liberopro_issue27}
------, ``Semantic (\_lan) perturbation never reaches the policy ({LIBERO-PRO}
  issue \#27),'' \url{https://github.com/Zxy-MLlab/LIBERO-PRO/issues/27}, 2026,
  accessed Nov.~2026.

\end{thebibliography}

\end{document}